\documentclass[letterpaper, 10 pt, conference]{ieeeconf}

\usepackage{times}
\usepackage{subfiles}
\usepackage{amsmath}
\usepackage{multicol}
\usepackage{float}
\usepackage{hyperref}
\usepackage{cleveref}
\usepackage{graphicx}
\usepackage{siunitx}

\usepackage{subcaption}
\usepackage{booktabs}
\usepackage{multirow}
\usepackage{tabularray}
\usepackage{colortbl}
\usepackage{pdfpages}

\usepackage[backend=biber,style=ieee]{biblatex}

\makeatletter
\newcommand\notsotiny{\@setfontsize\notsotiny\@vipt\@viipt}
\makeatother

\newcommand{\valc}[2]{\scalebox{0.95}{#1} \scalebox{.75}{$\pm$ #2}}

\newcommand{\val}[2]{\valc{#1}{#2}}

\DeclareMathOperator*{\argmax}{arg\,max}

\IEEEoverridecommandlockouts                              

\title{\LARGE \bf
Uncertainty-aware Active
Learning of NeRF-based Object Models for Robot Manipulators using Visual and Re-orientation Actions
}

\author{Saptarshi Dasgupta$^*$, Akshat Gupta$^*$, Shreshth Tuli and Rohan Paul \\
Indian Institute of Technology Delhi, India 
\hspace{2mm} {$^{\{*\}}$} denotes equal contribution
}

\begin{document}
\maketitle

\begin{abstract}
    Manipulating unseen objects is challenging without a 3D representation, as objects generally have occluded surfaces. 
This requires physical interaction with objects to build their internal representations. 
This paper presents an approach that enables a robot to rapidly learn the complete 3D model of a given object for manipulation in unfamiliar orientations. 
We use an ensemble of partially constructed NeRF models to quantify model uncertainty to determine the next action (a visual or re-orientation action) by optimizing informativeness and feasibility. 
Further, our approach determines \emph{when} and \emph{how} to grasp and re-orient an object given its partial NeRF model and re-estimates the object pose to rectify misalignments introduced during the interaction. 
Experiments with a simulated Franka Emika Robot Manipulator operating in a tabletop environment with benchmark objects demonstrate an improvement of (i) $14\%$ in visual reconstruction quality (PSNR), (ii) $20\%$ in the geometric/depth reconstruction of the object surface (F-score) and (iii) $71\%$ in the task success rate of manipulating objects \emph{a-priori} unseen orientations/stable configurations in the scene; over current methods.  
The project page can be found \href{https://actnerf.github.io/}{here}.
\end{abstract}
\renewcommand{\arraystretch}{0.9}
\section{Introduction}

We consider the problem of acquiring a 3D visual and geometric representation of an object for sequential robot manipulation tasks. In recent years, Neural Radiance Fields (NeRF) has emerged as a useful implicit representation that allows synthesis of novel views aiding in downstream planning, manipulation, and pose estimation tasks. Such a representation is acquired by collecting a set of views from known poses in the environment. The process for collecting such views is either in (i) batch mode~\cite{mildenhall2020nerf,yu2020pixelnerf,johari2022geonerf} by exhaustively collecting observations covering a region or (ii) actively by determining a set of informative views~\cite{lee2022uncertainty,lin2022active}.  
Although effective in rapidly constructing an object model, such approaches can only reconstruct the visible regions of the object, failing to model obscured parts such as the base, internal contents, and other occluded regions. The inability to accurately model the object owing to occlusions in the scene translates to poor manipulation ability for subsequent manipulation tasks.

\begin{figure}[t]
    \setlength{\belowcaptionskip}{-25pt}
     \centering
        \centering
        \includegraphics[width=0.5\textwidth]{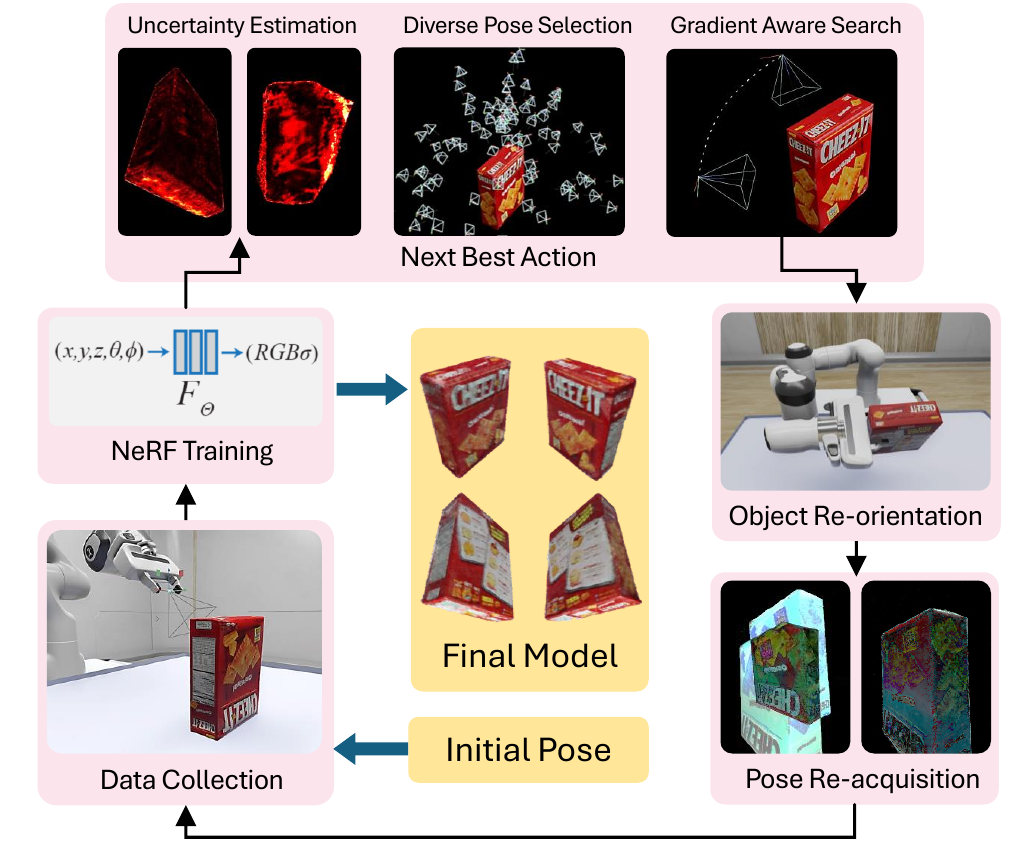}
        \caption{\textbf{Overview.} We present an active learning approach for building a NeRF model of an object that allows the robot to re-orient the object while collecting visual observations. Our approach is guided by a measure of model uncertainty in the partially constructed model, which is used to determine the most informative feasible action, aiding model construction. }
        \label{fig:intro_diagram}
     \hfill
\end{figure}

This work considers the possibility of \emph{directly interacting} via grasping, re-orientation, and stably releasing the object to expose previously \emph{unexposed} regions for subsequent model building. Fig. \ref{fig:intro_diagram} presents an overview of our model acquisition technique. 
Introducing physical interaction during model acquisition poses two key challenges. 
First, finding stable grasping points using a \emph{partially} built model is challenging due to depth uncertainty in unobserved or poorly observed regions. 
Second, re-orientation introduces uncertainty in the object's pose, affecting the incremental fusion of the radiance field arising from new observations. 
Further, as opposed to scene-based representations, we seek the ability to acquire object-centric radiance fields to support semantic tasks that may require sequential manipulation actions (e.g., clearing objects from a region). 

Overall, this paper makes the following contributions: 
\begin{itemize}
\item Leveraging vision foundation models to isolate the object of interest to disentangle its uncertainty from that of other background objects in the scene.
\item A search procedure that estimates the next most informative action (visual or re-orientation). The procedure relies on a coarse-to-fine optimization of the continuous viewing space incorporating (i) model uncertainty in the partially-built model (adapting~\cite{lin2022active}), (ii) motion costs, and (iii) kinematic constraints. 
\item An approach for grasping while accounting for the uncertainty in the partially constructed model and re-estimating the pose of the object after interaction for fusing the incrementally acquired model. 
\end{itemize}
Extensive evaluation with a simulated robot manipulator with benchmark objects shows improvements in the coverage and visual/geometric quality of the acquired model. 
Overall, this work takes a step in the direction of acquiring a rich  NeRF model of an object to support future robot manipulation tasks such as pick/place from arbitrary object configurations. 

\section{Related Work}
NeRF-based~\cite{mildenhall2020nerf} representations have been used in many robotics problems. DexNerf~\cite{Ichnowski2021DexNeRFUA}, and EvoNeRF~\cite{kerr2022evo} use NeRFs for modeling transparent objects that are difficult to represent with voxel-based methods. Adamkiewicz et al.~\cite{adamkiewicz_vision-only_2022} uses NeRFs to model the environment and synthesize trajectories for a quadrotor, while Driess et al.~\cite{2022-driess-compNerf} use NeRF for representing multi-object scenes and train graph neural networks to learn dynamics models. It is worth noting that while the aforementioned approaches utilize NeRF models for robotic tasks, they do not directly address the problem of determining the optimal viewpoints for constructing said NeRF models.

The concept of actively constructing a NeRF model has garnered attention in existing literature, closely intertwined with the next-best-view (NBV) problem, which entails identifying the optimal sensor location to maximize information acquisition about a given object or scene.
Traditional approaches for tackling the NBV problem include~\cite{krainin2011autonomous, isler2016information, daudelin2017adaptable}, who build volumetric 3D models through active learning. More recently, Lee et al.~\cite{lee2022uncertainty}, and NeU-NBV~\cite{jin2023neu} have delved into constructing implicit neural models by addressing the NBV problem within a robotic framework. Additionally, ActiveNeRF\cite{pan2022activenerf} and Lin et al. \cite{lin2022active} have approached the NBV problem purely from a visual perspective, without a robot manipulator. Central to these NBV techniques is characterizing \textit{model uncertainty} or the internal uncertainty estimates of the robot's own model. Several approaches have been proposed to quantify the uncertainty in NeRF models. S-NeRF~\cite{shen2021stochastic}, ProbNeRF~\cite{hoffman2023probnerf}, and ActiveNeRF~\cite{pan2022activenerf} integrate uncertainty prediction directly into the NeRF architecture. Lee et al.~\cite{lee2022uncertainty} models uncertainty as the entropy of the weight distribution along camera rays. Lin et al.~\cite{lin2022active} leverage variance in NeRF ensemble renderings for uncertainty quantification, while Sunderhauf et al.~\cite{sunderhauf2023density}  employ a combination of ensemble variance and termination probabilities along rays. \looseness = -1

Our work differs from the NBV approaches discussed above in two key aspects: firstly, by incorporating costs associated with each action and the robot's kinematics constraints, and secondly, by addressing the challenge of finding the next-best-view in the continuous SE(3) space while also permitting discrete actions through robot interactions, rather than focusing solely on selecting the best k images from a discrete set of (image, camera-pose) pairs.

\section{Background and Problem Setup}
\label{sec:background}
\subsection{NeRF-based Object Models}
Over the recent years, Neural Radiance Fields (NeRF)~\cite{mildenhall2020nerf} have gained prominence as an effective implicit neural representation technique for synthesizing novel views of a scene from a set of $N$ RGB images and their associated camera poses. NeRF employs a neural network to represent each scene, predicting both the volumetric density and view-dependent color for any given point within the scene. Specifically, the volumetric density $\sigma$ and RGB color $\mathbf{c}$ for each scene point are computed based on the parameters $\Theta$ of a Multilayer Perceptron (MLP), denoted by $F$. This MLP, is characterized by its input comprising the 3D position $\mathbf{x} = (x, y, z)$ and the viewing direction $\mathbf{d} = (d_x, d_y, d_z)$, outputs the ordered pair $(\sigma,\mathbf{c})$, collectively defining the scene's \emph{radiance field}. \looseness = -1

To render a novel view, NeRF traces camera rays for each pixel on the image plane, parameterized as $\mathbf{r}(t) = \mathbf{o} + t\mathbf{d}$, where $t \geq 0$, $\mathbf{o}$ represents the camera origin, and $\mathbf{d}$ is the unit vector in the direction of the ray. For each ray, $N$ points $\{\mathbf{r}_i = \mathbf{o} + t_i\mathbf{d}\}_{i=1}^N$ are sampled and processed by the MLP to obtain densities and colors. These are then integrated using volume rendering techniques (for further details, refer to \cite{mildenhall2020nerf}) to approximate the color $\boldsymbol{\hat{C}}(\mathbf{r})$, depth $\hat{D}(\mathbf{r})$, and opacity $\hat{O}(\mathbf{r})$ of each pixel. The NeRF model approximates these quantities using the Quadrature Rule~\cite{max1995optical}, expressed as follows:
\begin{align} 
    \boldsymbol{\hat{C}}(\mathbf{r}) &= \sum_{i=1}^N \alpha_i\mathbf{c}_i, \hat{D}(\mathbf{r}) = \sum_{i=1}^N  \alpha_i t_i, \hat{O}(\mathbf{r}) = \sum_{i=1}^N\alpha_i, \label{eq:nerf_estimations} \\
    \alpha_i &= \exp\left(-\sum_{j=1}^{i-1}\sigma_j \delta_j\right) \left(1-\exp(-\sigma_i \delta_i)\right), \label{eq:nerf_alphai_exp} 
\end{align}
where $\sigma_i$ and $\mathbf{c}_i$ denote the density and color predicted by the model at point $\mathbf{r}_i$ along ray $\mathbf{r}$, respectively, and $\delta_i = t_{i+1} - t_i$ represents the distance between adjacent samples along the ray.\looseness=-1

\begin{figure*}[t!]
    \setlength{\belowcaptionskip}{-20pt}
    \centering
        \centering
        \includegraphics[width=\textwidth]{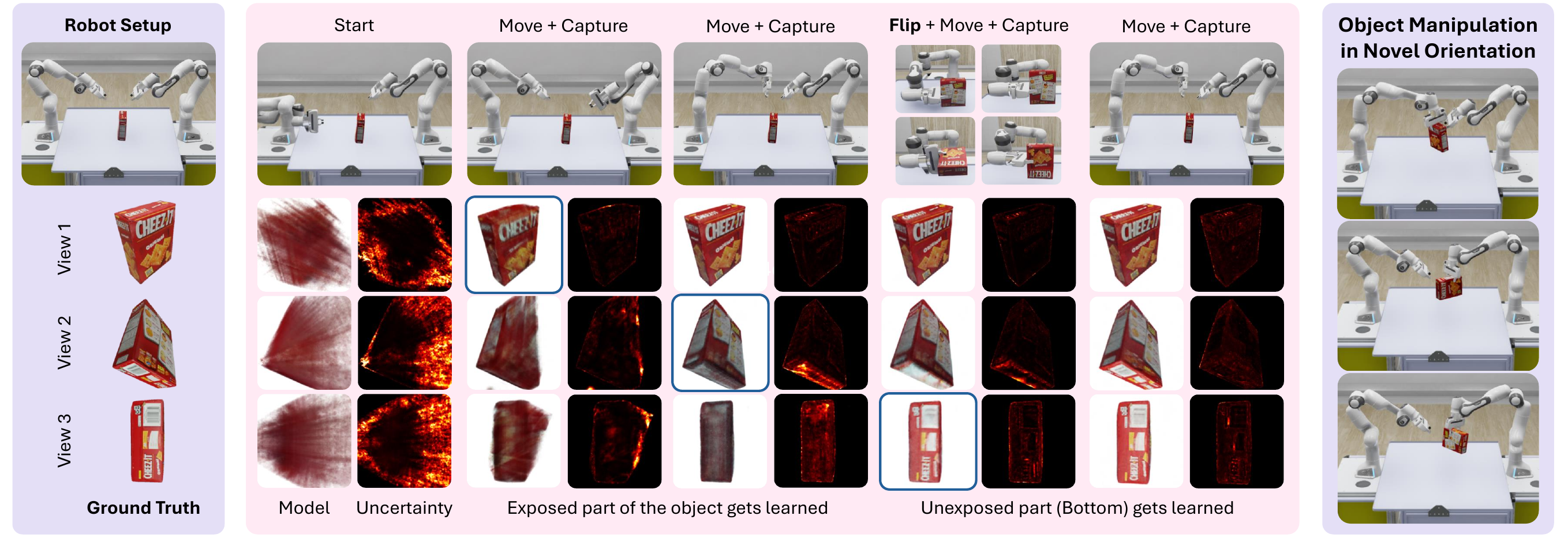}
        \caption{\textbf{Active Learning in Action. } We show the RGB images and uncertainty maps rendered from trained models during our active learning process. The GT images are shown for reference. We note from the figure that before flipping, the bottom surface of the object has high uncertainty, which only diminishes once we perform the flip and acquire information about the bottom surface. The robot then uses the acquired object model to manipulate the object in any orientation.}
        \label{fig:active_learning}
     \hfill
\end{figure*}
\subsection{Learning NeRF-based Object Models}
Our problem concerns a robot manipulator in a tabletop environment and an object placed near the table's center. The robot is tasked to acquire a 3D representation of the object, which can be subsequently leveraged to manipulate the object in any position and orientation. Let $A$ denote the robot's actions which include:  \verb|Move(|$p_i$\verb|)| which position the robot arm to $SE(3)$ pose $p_i$, \verb|Flip()|, which allows the robot to flip an object within its grasp using its object model, and \verb|Capture()|, where the robot acquires an image from the camera attached to the robot arm. Further, let $\Gamma(a)$ denote the cost of an action $a\in A$. 

The robot is required to execute a sequence of actions $A^* = (a_1, a_2, \ldots, a_n)$, where each $a_i$ represents a specific combination of actions from $A$. After executing each action $a_i$, the robot applies a capture function $Capture()$ action to obtain an image. The collected images $I^* = (I_1, I_2, \ldots, I_n)$ are then used to train a NeRF model $F_\theta$. Given a partially trained model $F_{\Theta_{k-1}}$, based on images $i_1, i_2, \ldots, i_{k-1}$, the goal is to identify the next action $a_k$ that enables the robot to capture an image from a viewpoint where the model exhibits the highest uncertainty, while also minimizing the associated action cost $\Gamma(a_k)$. 

\section{Technical Approach}
Our approach for active learning of NeRF-based object models consists of (i) estimating model uncertainty for a partially-built model, (ii) determining the next informative and feasible action and (iii) incorporating object re-orientation and pose re-acquistion. These modules are detailed in this section (see Fig.~\ref{fig:active_learning}).    
Formally, we express the aforementioned objective as optimizing the following: 
\begin{equation}
a_k = \argmax_{a} \left[ U(F_{\Theta_{k-1}}, p) - \lambda \Gamma(a) \right],
\end{equation}
where, $p$ represents the 6 degrees of freedom (DoF) pose achieved by the robotic arm upon executing action $a$, and $U(F_{\Theta_{k-1}}, p)$ quantifies the uncertainty in the model from pose $p$.  The objective can be equivalently expressed as minimizing the loss function $L(a)$, defined as:
\begin{equation} \label{eq:loss_defn}
    L(a) = \lambda \Gamma(a) - U(F_{\Theta_{k-1}}, p)
\end{equation}

\subsection{Estimating Model Uncertainty} 
\label{sec:model_uncertainty}
\begin{figure}
    \centering
    \setlength{\belowcaptionskip}{-10pt}
    \includegraphics[width=0.8\linewidth]{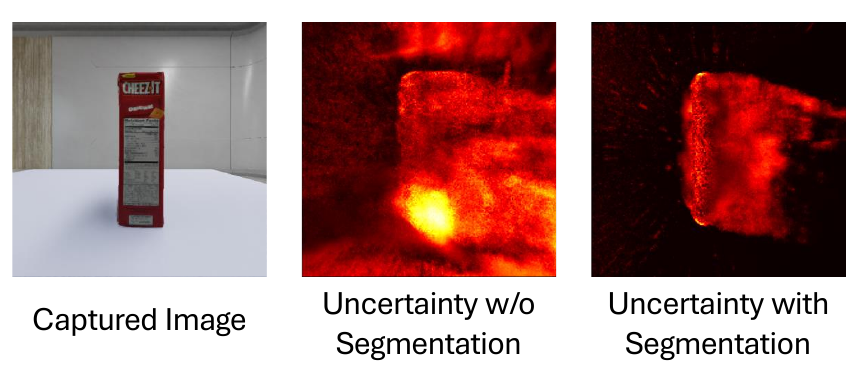}    
    \caption{\textbf{Necessity of Segmentation.} Uncertainty heatmaps of NeRF models trained on segmented v/s original images}
    \label{fig:unc_segment}
\end{figure}

As discussed in Section~\ref{sec:background}, quantifying the uncertainty present in a partial NeRF model from a given pose is crucial for our approach. Following the methodology proposed by Lin et al.~\cite{lin2022active}, we employ an ensemble-based strategy to measure this uncertainty. Specifically, we train $M$ NeRF models using the same set of images but initialize each model with distinct weights sampled from a Xavier uniform distribution. By rendering images from these $M$ models for any selected camera pose, we calculate the total variance across the RGB color channels and produce an uncertainty heatmap (see Fig.~\ref{fig:unc_segment}).

The overall uncertainty for a given pose is determined by aggregating the uncertainties of individual pixels within the rendered image. Therefore, the uncertainty associated with a pixel corresponding to ray $\mathbf{r}$ is defined as the variance of the estimated colors $\boldsymbol{\hat{C}}_{i}(\mathbf{r})$, calculated as follows:
\begin{align} \label{eq:unc_defn}
\sigma^2(\mathbf{r}) &= \frac{1}{M} \sum_{k=1}^{M} \| \boldsymbol{\mu}(\mathbf{r}) - \boldsymbol{\hat{C}}_{k}(\mathbf{r}) \|^2, \text{where}
\\
\boldsymbol{\mu}(\mathbf{r}) &= \frac{1}{M} \sum_{k=1}^{M} \boldsymbol{\hat{C}}_{k}(\mathbf{r}),
\end{align}
and $\boldsymbol{\mu}(\mathbf{r})$ and $\boldsymbol{\hat{C}}_{i}(\mathbf{r})$ are vectors representing the RGB color channels. Here, $M$ denotes the total number of models in the NeRF ensemble. Using the expression for $\sigma^2(\mathbf{r})$, the uncertainty for a pose $p$ can be quantified as the sum of the uncertainties for all rays emanating from $p$:
\begin{equation}
    \label{eq:unc_wrt_rays}
    U(F_{\Theta_{k-1}}, p) = \sum_{\mathbf{r} \in \text{Rays}(p)}{\sigma^2(\mathbf{r})}.
\end{equation}

\begin{figure*}[t!]
     \centering
        \centering
        \setlength{\belowcaptionskip}{-20pt}
        \includegraphics[width=\textwidth]{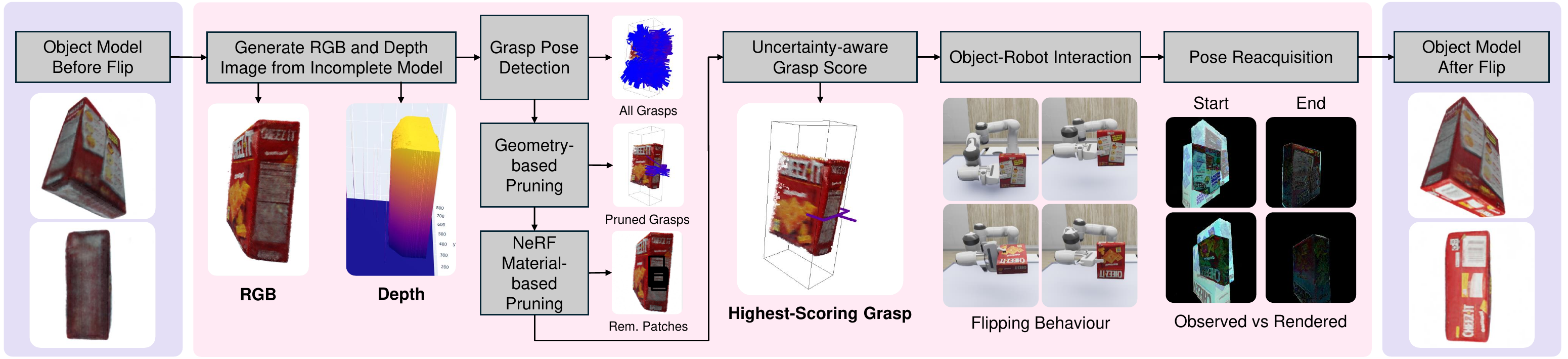}
        \caption{\textbf{Object Re-orientation Approach.} First, the RGB and Depth images are rendered from the object's current NeRF model. Using these, AnyGrasp detects potential grasps, which are then pruned based on the geometry of the generated point cloud and NeRF's material density on grasp patches. The best grasp is selected from the remaining using our uncertainty-aware grasp score. The robot executes the chosen grasp to re-orient the object, and the modified iNeRF is employed to re-acquire the object's pose in its new orientation. We show the quality of the object models before and after the flip. The post-flip model is obtained by capturing images in a re-oriented position and adding them to the training dataset. }
        \label{fig:flip_process}
     \hfill
\end{figure*}

Note that creating a 3D representation encompassing the entire scene results in an estimated uncertainty that reflects both the object of interest and the surrounding environment. Consequently, employing an uncertainty-based next-best-view (NBV) strategy under such conditions inadvertently optimizes for the reduction of background uncertainty as well, which diverges from our primary objective. Moreover, this method proves ineffective in cluttered scenes populated with multiple objects. Our aim is to isolate and enhance the uncertainty associated exclusively with the object of interest. To this end, we employ Grounded-SAM~\cite{ren2024grounded}, a technique that utilizes textual prompts to generate object masks through the integration of Grounding DINO~\cite{liu2023grounding} and SAM~\cite{kirillov2023segany}, facilitating the training of NeRF models on segmented images. This approach provides a more accurate assessment of model uncertainty from the perspective of the object of interest (refer to Fig.~\ref{fig:unc_segment}).

S{\"u}nderhauf et al.~\cite{sunderhauf2023density} argue that RGB uncertainty does not adequately represent the model's epistemic uncertainty, particularly in relation to scene elements that remain unobserved during training. They propose quantifying epistemic uncertainty via the aggregation of termination probabilities for points sampled along each ray, noting that uncertainty peaks when rays fail to intersect with the scene. However, this method is not applicable to our scenario, as our focus lies on quantifying object-specific uncertainty rather than that of the entire scene. Rays that do not intersect with the object contribute to an increased epistemic uncertainty, particularly for camera views distant from or oriented away from the object. Identifying and filtering out rays that do not intersect with the object of interest is a much harder problem with an apriori unknown object model. To circumvent this, we use RGB uncertainty as a proxy metric that effectively indicates heightened uncertainty in views of the object that have not been previously observed. Additionally, we conduct ablation studies comparing our approach with a modified version of their uncertainty measure, as detailed in Section~\ref{sec:ablations} to further highlight that RGB uncertainty is more amenable for robotic manipulation scenarios.

\subsection{Uncertainty-guided Next Action Selection}
\label{sec:next_action_selection}
We now tackle the challenge of identifying the next best action within the context of our active learning framework, given the current training dataset of the NeRF model and the robot's present pose. This task is formalized as minimizing the objective function $L(a)$, as defined in (\ref{eq:loss_defn}), where $U(F_{\Theta_{k-1}}, p)$ is articulated in (\ref{eq:unc_wrt_rays}). A notable issue arises due to the significant variance in uncertainty values across different NeRF models, even when trained on disparate images of the same object. To address this and standardize the selection of the $\lambda$ parameter across all models, we normalize the uncertainty derived from (\ref{eq:unc_wrt_rays}) by the model's mean uncertainty, calculated over a set of poses randomly sampled from a uniform spherical distribution, ensuring the uncertainty prediction is \textit{model-agnostic}.

Initially, we consider a simplified scenario where only \verb|Move()| actions are permissible. This is because the model is not exposed to enough images to build a reasonable 3D representation required for grasping and consequently flipping. In this case, the minimization variable $a$ in (\ref{eq:loss_defn}) is substituted with $p \in SE(3)$, representing the 6-DoF pose of the camera affixed to the robot's end-effector. The designated action for a pose $p$ corresponds to maneuvering the end-effector to position the camera at $p$. To circumvent the limitations of naive gradient descent approaches, which falter due to the presence of numerous local minima within the objective function, we propose a bi-level optimization strategy. The primary level involves selecting a \textit{sparse} and \textit{diverse} subset of $k$ candidate poses from $n$ randomly sampled poses, all oriented towards the workspace's center. Subsequently, at the secondary level, we execute a gradient descent search from each candidate pose, mitigating the risk of converging to suboptimal local minima. The final solution is determined by selecting the candidate pose from the second level that yields the lowest objective function value.

Subsequently, in scenarios where \verb|Flip()| actions are also considered, the problem is decomposed into two subproblems: 1) Identifying the optimal action assuming no flip action is permitted, and 2) Adjusting the coordinate axes to simulate a flip action and determining the optimal subsequent action. The cost associated with \verb|Flip()| is accounted for exclusively in the second subproblem. The ultimate optimal action is selected based on the lower value of $L(a)$ obtained from these subproblems. 

Our methodology accommodates any form of action cost $\Gamma(a)$ specified in (\ref{eq:loss_defn}). 
For our experiments, we define $\Gamma(a)$ for a \verb|Move()| action, which transitions the end-effector from $(r_1, q_1)$ to $(r_2, q_2)$ (with $r$ indicating position and $q$ representing rotation in quaternion form), as follows:

\begin{equation}
    \label{eq:action_cost_defn}
    \Gamma(a) = \alpha_1 (1 - d(q_1, q_2))  + \alpha_2 d(r_1, r_2),
\end{equation}
whereas, for a \verb|Flip()| action, we set $\Gamma(a) = \alpha_3$. The cumulative cost $\Gamma$ for a sequence of actions is the sum of the costs for individual actions, where $\alpha_i$ are adjustable based on the relative importance of each action cost component.

\begin{table*}[t]
\captionsetup{font=small}
\caption{Comparison of our method with ActiveNeRF and other baselines.  All the baselines include \texttt{Flip()} action as detailed in section \ref{baselines_metrics} and are trained with segmented images. F-score* represents the F-score values multiplied by 10. As evident from the tables, our approach outperforms other methods by a significant margin.}
\setlength{\abovecaptionskip}{5pt}
\setlength{\belowcaptionskip}{5pt}

\begin{subtable}{\textwidth}
\centering
\captionsetup{font=footnotesize}
\bgroup
\setlength\tabcolsep{3.5pt}
\begin{tabular}{@{}lcccccccccccc@{}}
\toprule
\multicolumn{1}{l}{\multirow{2}[2]{*}{Method}} & \multicolumn{2}{c}{Basket} & \multicolumn{2}{c}{Cheezit Box} & \multicolumn{2}{c}{Mug} & \multicolumn{2}{c}{Rubik's Cube} & \multicolumn{2}{c}{Spam Can} & \multicolumn{2}{c}{\textbf{Total}} \\
\cmidrule(lr){2-11}\cmidrule(lr){12-13}  & PSNR$\uparrow$  & F-score*$\uparrow$ & PSNR$\uparrow$  & F-score*$\uparrow$ & PSNR$\uparrow$  & F-score*$\uparrow$ & PSNR$\uparrow$  & F-score*$\uparrow$ & PSNR$\uparrow$  & F-score*$\uparrow$ & PSNR$\uparrow$  & F-score*$\uparrow$ \\

\midrule

\multicolumn{13}{c}{Model quality after 20 iterations \textbf{without} grasping}\\

\midrule

Ours   & \textbf{\val{17.2}{0.1}} & \textbf{\val{4.2}{0.5}} & \textbf{\val{21.8}{0.2}} & \textbf{\val{4.3}{0.2}} & \val{26.3}{0.1} & \val{6.5}{0.2} & \textbf{\val{30.3}{0.3}} & \val{3.8}{0.1} & \val{23.9}{0.6} & \val{4.1}{0.3} & \textbf{\val{23.9}{0.1}} & \textbf{\val{4.6}{0.1}} \\
Random & \val{17.0}{0.2} & \val{4.0}{0.6} & \val{21.4}{0.4} & \val{4.0}{0.4} & \textbf{\val{26.5}{0.2}} & \textbf{\val{6.6}{0.1}} & \val{29.5}{0.2} & \textbf{\val{4.1}{0.3}} & \textbf{\val{24.7}{0.2}} & \textbf{\val{4.4}{0.0}} & \val{23.8}{0.1} & \val{4.6}{0.2} \\
Furthest & \val{17.1}{0.1} & \val{3.0}{0.2} & \val{21.2}{0.1} & \val{4.2}{0.2} & \val{26.5}{0.3} & \val{5.8}{0.2} & \val{28.6}{0.5} & \val{3.6}{0.2} & \val{23.9}{0.3} & \val{4.0}{0.2} & \val{23.5}{0.1} & \val{4.1}{0.1} \\
Active \cite{pan2022activenerf} & \val{15.5}{0.5} & \val{2.6}{0.2} & \val{19.0}{0.1} & \val{2.9}{0.3} & \val{24.8}{0.9} & \val{4.4}{0.1} & \val{27.0}{0.6} & \val{3.4}{0.3} & \val{19.7}{1.3} & \val{3.1}{0.1} & \val{21.2}{0.4} & \val{3.3}{0.1} \\

\midrule

\multicolumn{13}{c}{Model quality after 20 iterations \textbf{with} grasping}\\
\midrule
Ours   & \textbf{\val{16.9}{0.1}} & \val{3.4}{0.3} & \textbf{\val{21.3}{0.3}} & \textbf{\val{4.0}{0.2}} & \textbf{\val{26.9}{0.2}} & \textbf{\val{5.9}{0.1}} & \textbf{\val{31.6}{0.5}} & \textbf{\val{4.0}{0.3}} & \textbf{\val{23.4}{0.5}} & \textbf{\val{3.8}{0.1}} & \textbf{\val{24.0}{0.2}} & \textbf{\val{4.2}{0.1}} \\
Random & \val{16.0}{0.6} & \textbf{\val{3.5}{0.8}} & \val{20.9}{1.0} & \val{3.9}{0.4} & \val{22.2}{0.8} & \val{4.6}{0.2} & \val{29.1}{0.9} & \val{3.5}{0.5} & \val{22.8}{1.8} & \val{3.4}{0.4} & \val{22.2}{0.5} & \val{3.8}{0.2} \\
Furthest & \val{16.5}{0.2} & \val{3.1}{0.2} & \val{19.8}{0.7} & \val{3.7}{0.4} & \val{24.1}{0.3} & \val{4.8}{0.7} & \val{27.8}{0.9} & \val{3.8}{0.4} & \val{21.1}{1.5} & \val{3.3}{0.3} & \val{21.9}{0.4} & \val{3.7}{0.2} \\
Active \cite{pan2022activenerf} & \val{16.1}{0.2} & \val{2.6}{0.2} & \val{19.2}{0.2} & \val{2.5}{0.9} & \val{23.6}{3.2} & \val{5.5}{0.3} & \val{27.6}{0.8} & \val{3.9}{0.6} & \val{22.1}{0.2} & \val{3.2}{0.1} & \val{21.7}{0.7} & \val{3.5}{0.2} \\

\midrule

\multicolumn{13}{c}{Model quality attained given a cost budget of 2 \textbf{without} grasping}\\
\midrule
Ours   & \textbf{\val{17.1}{0.1}} & \textbf{\val{3.9}{0.3}} & \textbf{\val{21.4}{0.2}} & \textbf{\val{4.3}{0.1}} & \textbf{\val{26.0}{0.3}} & \textbf{\val{5.8}{0.5}} & \textbf{\val{29.8}{0.7}} & \textbf{\val{4.5}{0.2}} & \textbf{\val{23.7}{0.4}} & \textbf{\val{4.1}{0.2}} & \textbf{\val{23.6}{0.2}} & \textbf{\val{4.5}{0.1}} \\
Random & \val{17.0}{0.2} & \val{3.8}{0.4} & \val{19.3}{1.1} & \val{3.7}{0.4} & \val{25.2}{0.9} & \val{5.7}{0.7} & \val{28.6}{0.6} & \val{3.8}{0.1} & \val{23.3}{1.4} & \val{3.6}{0.3} & \val{22.7}{0.4} & \val{4.1}{0.2} \\
Furthest & \val{16.4}{0.4} & \val{2.6}{0.1} & \val{19.2}{0.2} & \val{3.5}{0.1} & \val{24.7}{1.1} & \val{4.4}{0.1} & \val{26.6}{1.2} & \val{4.1}{0.1} & \val{22.6}{0.4} & \val{3.9}{0.1} & \val{21.9}{0.3} & \val{3.7}{0.0} \\
Active \cite{pan2022activenerf} & \val{15.4}{0.4} & \val{3.1}{0.1} & \val{18.4}{0.2} & \val{3.1}{0.1} & \val{24.1}{0.4} & \val{4.1}{0.2} & \val{26.3}{0.6} & \val{3.4}{0.3} & \val{19.5}{1.7} & \val{3.6}{0.1} & \val{20.7}{0.4} & \val{3.5}{0.1} \\
\midrule

\multicolumn{13}{c}{Model quality attained given a cost budget of 2 \textbf{with} grasping}\\

\midrule
Ours   & \textbf{\val{16.9}{0.1}} & \textbf{\val{3.3}{0.4}} & \textbf{\val{21.2}{0.2}} & \textbf{\val{3.9}{0.2}} & \textbf{\val{26.8}{0.4}} & \textbf{\val{5.8}{0.3}} & \textbf{\val{30.9}{0.7}} & \textbf{\val{4.0}{0.3}} & \textbf{\val{23.7}{0.6}} & \textbf{\val{3.8}{0.2}} & \textbf{\val{23.9}{0.2}} & \textbf{\val{4.2}{0.1}} \\
Random & \val{16.0}{0.3} & \val{3.0}{0.3} & \val{20.9}{1.0} & \val{3.8}{0.2} & \val{22.3}{0.8} & \val{4.4}{0.2} & \val{28.5}{1.3} & \val{3.6}{0.3} & \val{22.8}{1.9} & \val{3.4}{0.4} & \val{22.1}{0.5} & \val{3.6}{0.1} \\
Furthest & \val{16.5}{0.2} & \val{1.5}{0.8} & \val{19.9}{0.2} & \val{3.6}{0.2} & \val{23.5}{0.4} & \val{4.3}{0.3} & \val{27.7}{0.9} & \val{3.5}{0.4} & \val{22.1}{1.7} & \val{3.3}{0.3} & \val{21.9}{0.4} & \val{3.2}{0.2} \\
Active \cite{pan2022activenerf} & \val{15.8}{0.2} & \val{3.3}{0.2} & \val{19.2}{0.3} & \val{3.2}{0.1} & \val{21.9}{2.0} & \val{5.1}{0.2} & \val{26.4}{0.8} & \val{2.7}{1.0} & \val{21.6}{0.5} & \val{3.3}{0.2} & \val{21.0}{0.8} & \val{3.5}{0.2} \\


\bottomrule
\end{tabular}%
\egroup
\end{subtable}

\label{tab:metrics_experiments}
\end{table*}

\subsection{Object Re-orientation during Model Acquisition}
\label{sec:reorientation}
In the subsequent phase of our methodology, we delve into the estimation of the grasp pose (see Fig. \ref{fig:flip_process}).
To compute the optimal lateral grasp pose based on the currently available partial NeRF model, we employ AnyGrasp~\cite{fang2023anygrasp}. The quality of the selected grasp pose significantly influences the decision-making process of our next-best-action algorithm, particularly in determining the possibility of a flip action in the ensuing iteration. AnyGrasp operates by processing depth images, from which it generates a collection of grasp pose and grasp confidence pairings. However, our empirical observations reveal that the confidence scores produced by AnyGrasp are not directly applicable for selecting grasp poses as it is trained on RGBD images obtained from depth cameras, which contrasts with our utilization of partial object models that may include extraneous geometric features.  We use our partially trained NeRF model to generate a depth image for a horizontal grasp. We then generate all candidate grasp poses using AnyGrasp. We prune grasp poses based on the following criteria: (i) Distance from center of the point cloud, (ii) Grasp angle w.r.t. surface normal, and (iii) Average opacity $\hat{O}(\textbf{r})$ (\ref{eq:nerf_estimations}) of the grasp patch. We then score to each grasp pose, to select the most suitable grasp. Our grasp score is defined as follows
\begin{equation} 
\label{eq:grasp_score}
G_s = \frac{1 - \theta}{U_d}
\end{equation}
where $\theta$ is the angle between the grasp pose and surface normal. $\theta$ should be minimized as grasping from a non-lateral grasp increases the probability of the object toppling during or after the flip. ${U_d}$ is the variance of rendered depths summed over the rays of the grasp patch. Its computation is similar to that of (\ref{eq:unc_defn}), and (\ref{eq:unc_wrt_rays}) with predicted depth $\hat{D}(\textbf{r})$~(\ref{eq:nerf_estimations}) being used instead of color. We minimize $U_d$ to be certain about the location of the object surface in 3D space near the grasp pose.  This approach ensures that the grasp poses chosen are not only theoretically viable according to AnyGrasp's criteria but also practically applicable within the constraints and current state of our partial object models.

\begin{figure*}
    \centering
    \begin{subfigure}[b]{0.19\textwidth}
        \includegraphics[width=\textwidth]{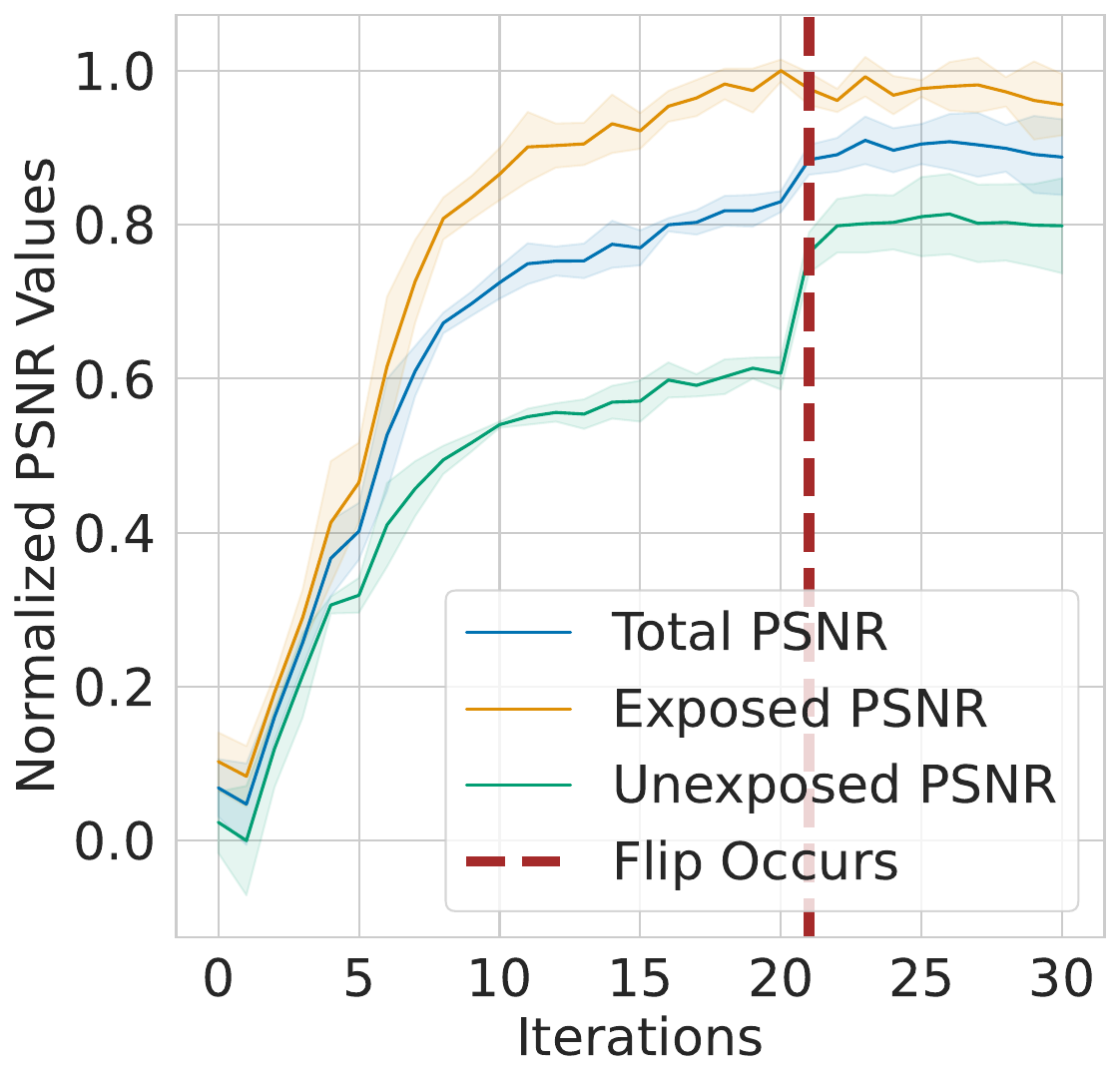}
        \caption{Cheezit Box}
        \label{fig:rq2_psnr_iterations_cheezit}
    \end{subfigure}
    \begin{subfigure}[b]{0.19\textwidth}
        \includegraphics[width=\textwidth]{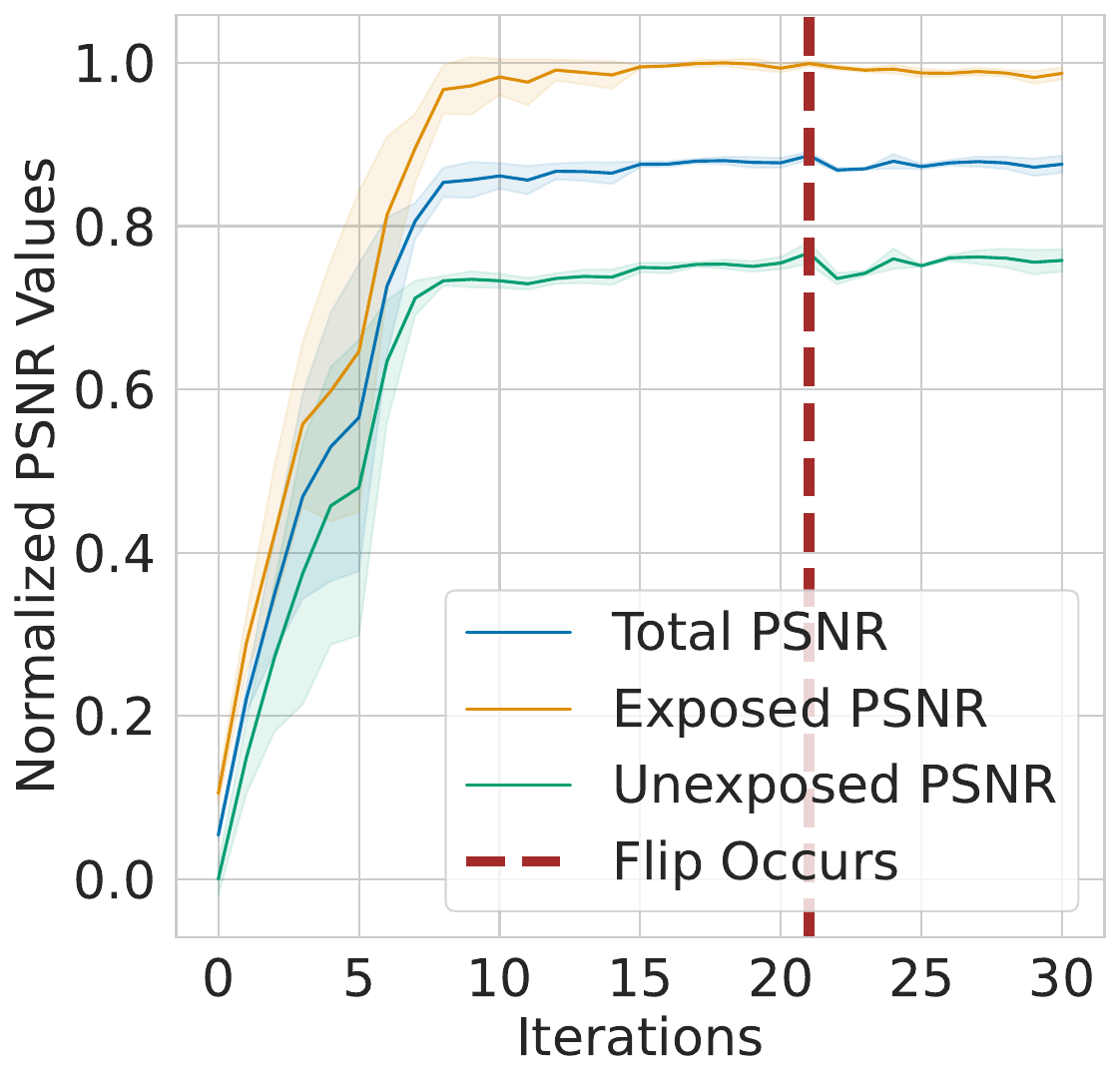}
        \caption{Basket}
        \label{fig:rq2_psnr_iterations_basket}
    \end{subfigure}
    \begin{subfigure}[b]{0.19\textwidth}
        \includegraphics[width=\textwidth]{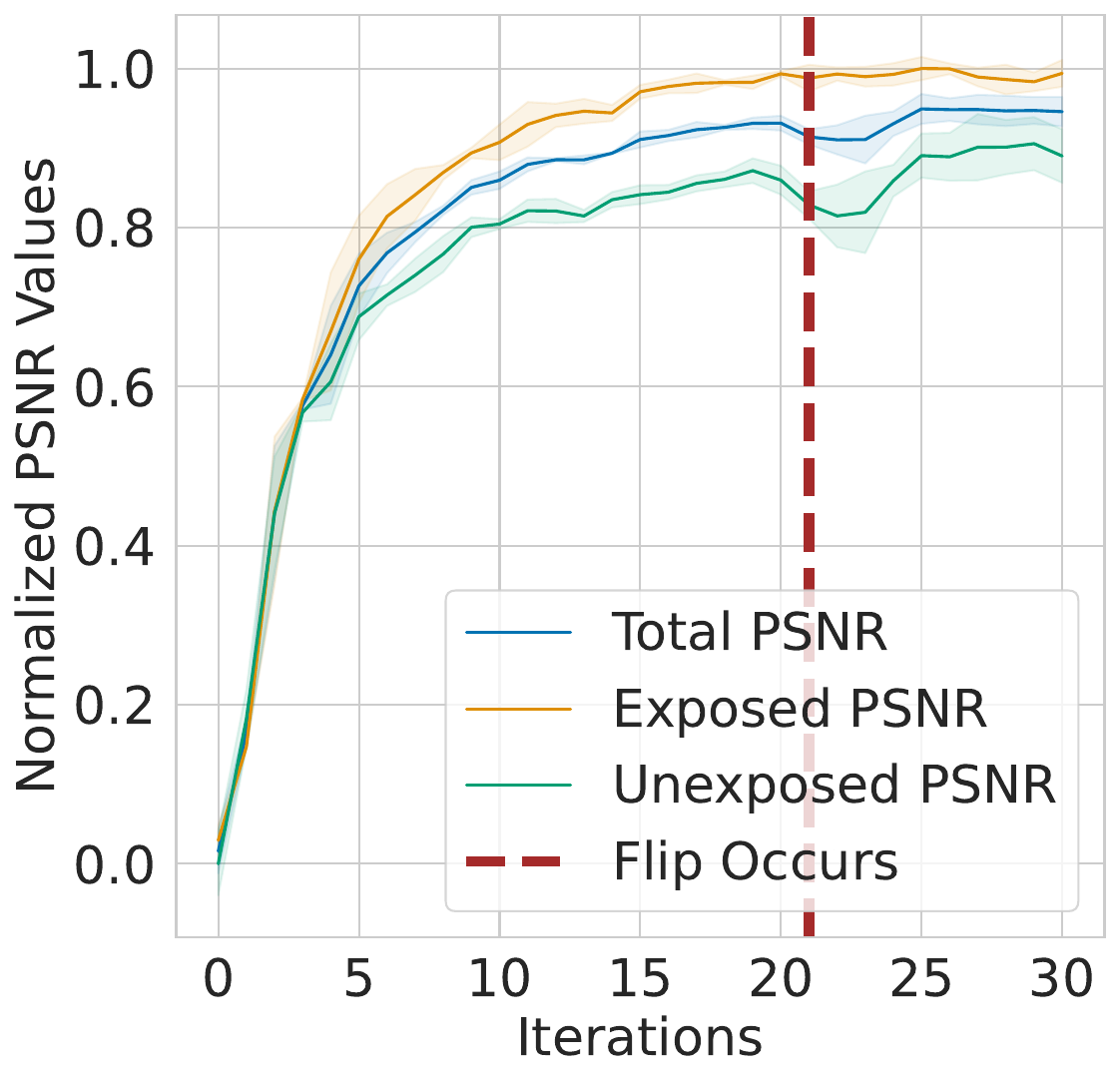}
        \caption{Mug}
        \label{fig:rq2_psnr_iterations_mug}
    \end{subfigure}
    \begin{subfigure}[b]{0.19\textwidth}
        \includegraphics[width=\textwidth]{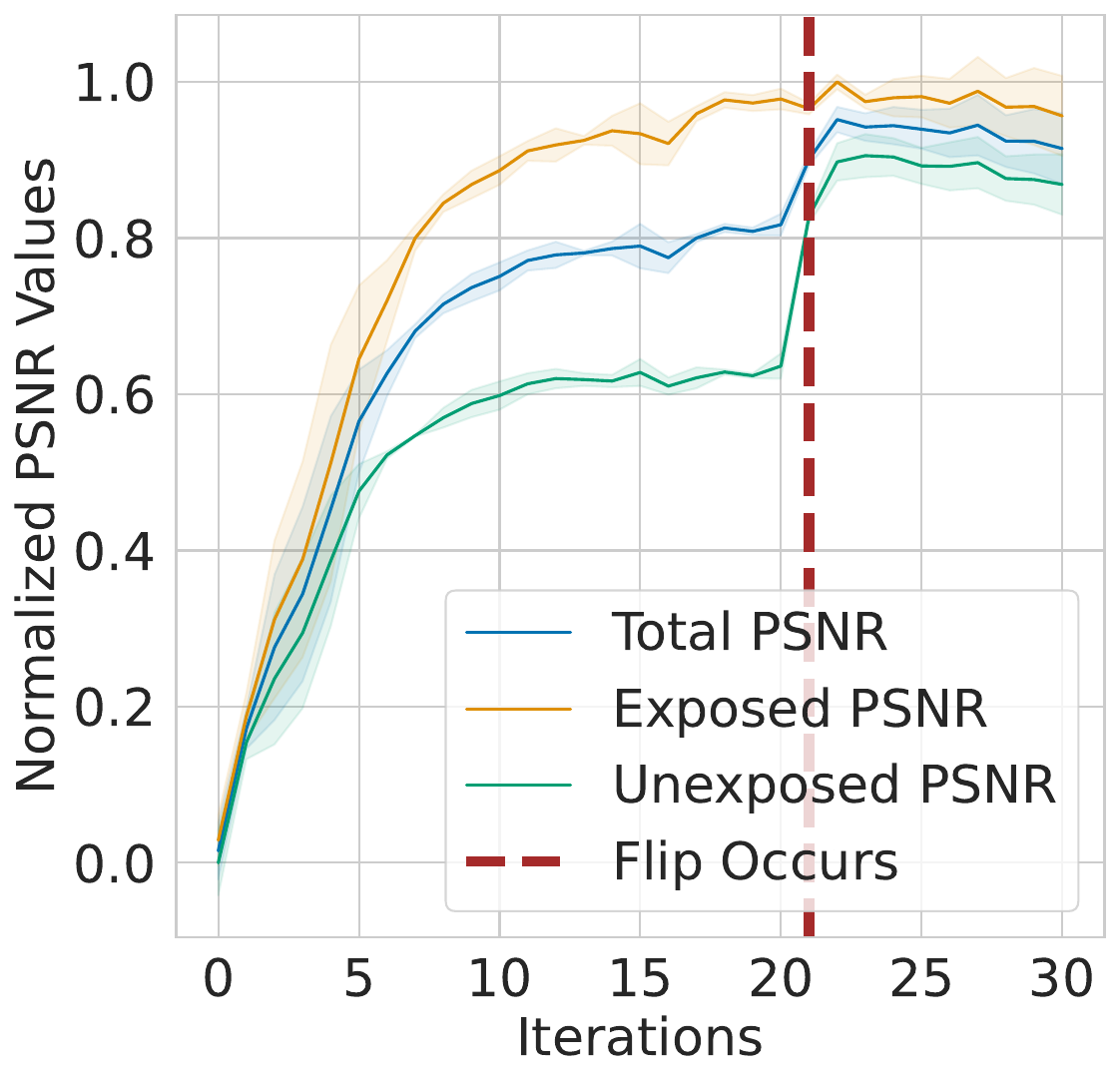}
        \caption{Rubik's Cube}
        \label{fig:rq2_psnr_iterations_rubik}
    \end{subfigure}
    \begin{subfigure}[b]{0.19\textwidth}
        \includegraphics[width=\textwidth]{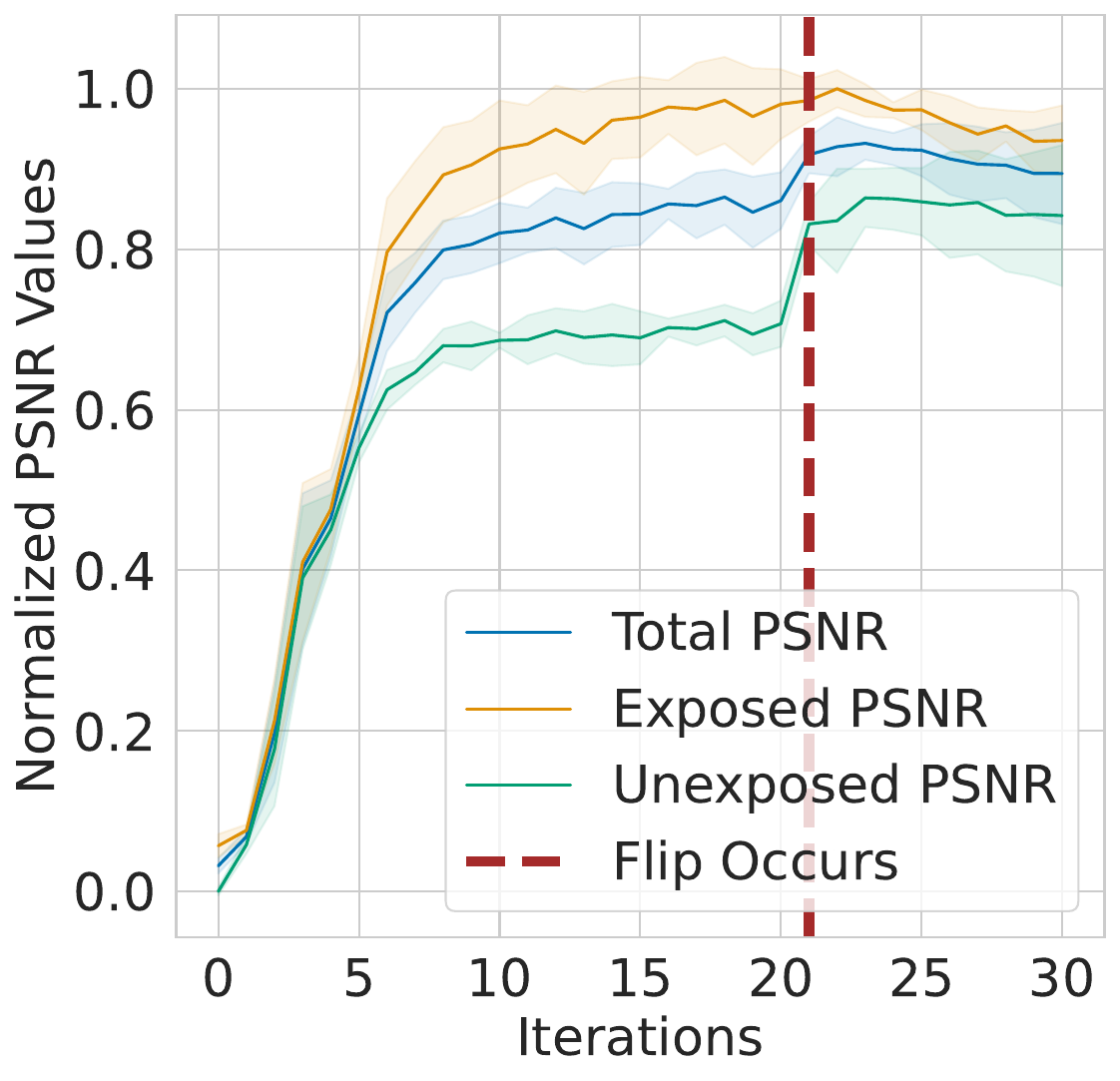}
        \caption{Spam Can}
        \label{fig:rq2_psnr_iterations_spam}
    \end{subfigure}
    \setlength{\belowcaptionskip}{-10pt}
    \caption{\textbf{Effect of Flip on Model Quality}.We demonstrate the impact of flipping objects on model quality (PSNR). The dashed line indicates the iteration at which the object is flipped. The PSNR is shown for the \textbf{Exposed} and \textbf{Unexposed} subsets of the validation set, representing camera poses above and below the object center, respectively. In most cases, the Exposed PSNR remains almost constant, while the Unexposed PSNR shows a significant increase, leading to an overall improvement in total PSNR. PSNR values are min-max normalized in each plot.}
    \label{fig:rq2_psnr_iterations}
\end{figure*}

\subsection{Pose Re-acquisition for Model Unification}
\label{sec:reacquisition}

The stochastic nature of robotic actions necessitates the recovery of an object's pose following the execution of a \verb|Flip()| action. To address this, we employ a methodology inspired by iNeRF~\cite{yen2021inerf}. Subsequent to the interaction, the robot captures an RGB image, the pose of which is ascertainable through the robot's forward kinematics. The alteration in the object's pose due to the interaction, however, results in discrepancies between the newly captured image and what NeRF would render from the same camera position. To reconcile these differences, we optimize for the camera pose that minimizes the Sum of Squared Differences (\textbf{SSD}) between the captured image and NeRF's predicted image, thereby enabling an estimation of the object's post-interaction pose.

Our experimentation reveals that the precision of pose estimation via the original iNeRF framework does not meet the requisite standards for eliminating the discrepancies in the data collected before and after re-orientation. Consequently, we introduce two significant enhancements to the conventional iNeRF approach. Firstly, diverging from iNeRF's gradient-based search methodology, we adopt non-gradient-based optimization techniques, which have demonstrated superior performance in accurately recovering object poses. Specifically, we combine three distinct optimization strategies: Nelder-Mead~\cite{nelder1965simplex}, COBYLA~\cite{powell1994direct,powell1998direct}, and Powell's method~\cite{powell1964efficient}. The most accurate pose estimation from among these methods is selected based on the lowest \textbf{SSD} score. Secondly, in lieu of relying on a single image for pose estimation, we utilize multiple images to enhance the robustness. The optimization process is thus aimed at minimizing the cumulative \textbf{SSD} across all pairs of captured and NeRF-predicted images. This multi-image strategy bolsters the accuracy of our pose estimation, ensuring a more reliable recovery of the object's pose post-interaction.

\begin{figure}[t]
    \setlength{\belowcaptionskip}{-15pt}
    \centering
    \includegraphics[width=\linewidth]{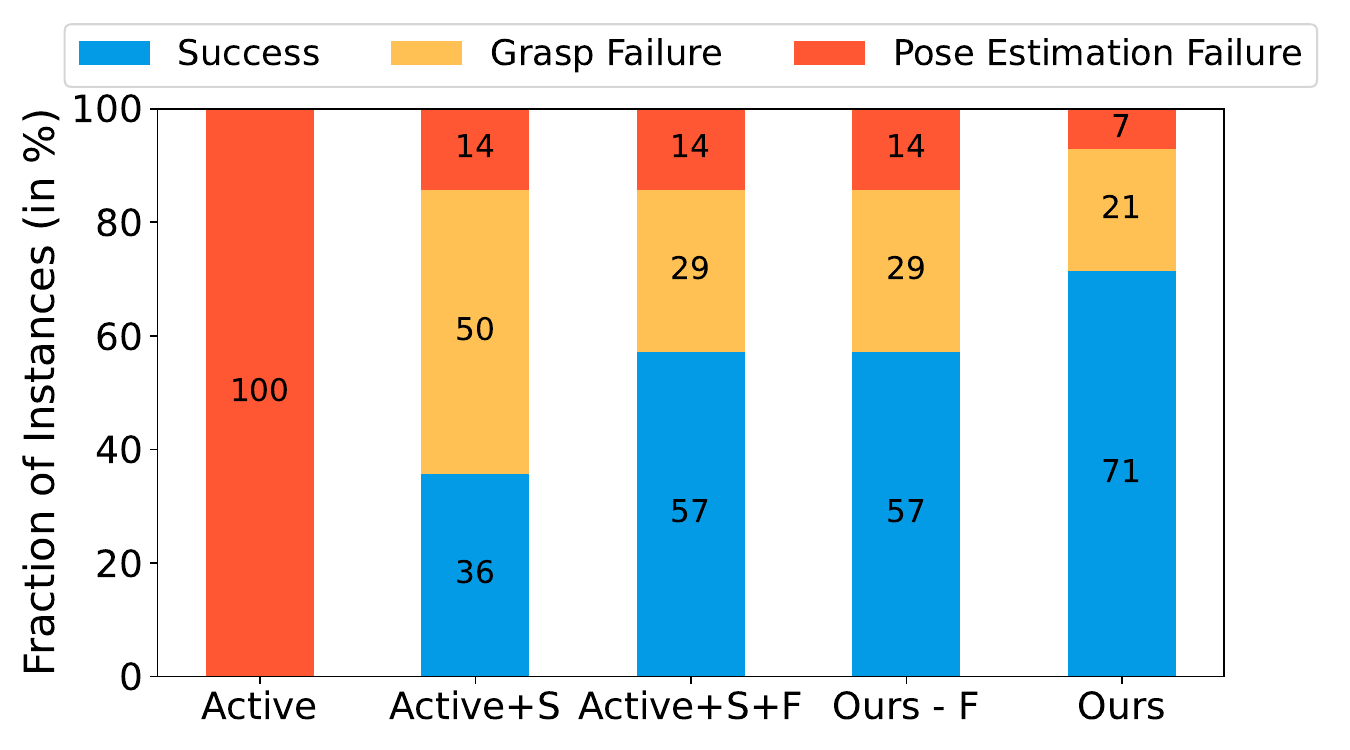}
    \caption{\textbf{Grasping Performance Analysis.} Analysis of grasp task success rate and failure scenarios. \textbf{Active} denotes vanilla ActiveNeRF trained on captured images without segmentation and with no \texttt{Flip()} action. \textbf{S} denotes \textit{object segmentation}, \textbf{F} denotes possibility of \texttt{Flip()} action.}
    \label{fig:task_based}
\end{figure}
\section{Evaluation Setup}  \label{evaluation_setup}
\subsection{Simulation Environment and Dataset}
Our experiments consider a table as a workspace with two Franka Emika robotic arms, situated at opposite ends. This dual-arm setup is necessitated by the limitations of a single arm's reach and inability to capture images covering the entirety of an object's surface, especially areas directly opposite the arm. To simulate a realistic environment conducive to our active learning endeavors, we employ Nvidia's \textit{Isaac Sim} simulator. We curate object models from the YCB dataset~\cite{calli2017yale}, focusing on objects amenable to lateral grasping and flipping. The dataset consists of five objects as shown in Fig \ref{fig:qual_res}.\looseness=-1

\subsection{Baselines and Metrics} \label{baselines_metrics}
Our evaluation framework benchmarks the proposed active learning strategy against three baselines to ensure a rigorous comparison: (i) \textit{Random View}, where subsequent views are randomly selected, (ii) \textit{Next Furthest View}, selecting the next view to maximize the cumulative distance from existing training views, and (iii) \textit{ActiveNeRF}~\cite{pan2022activenerf}, implemented via the Kaolin-Wisp~\cite{takikawa2022kaolin} framework. 
We integrated a \verb|Flip()| action in these baselines to align them with our framework.
Specifically, for (i) and (ii), a \verb|Flip()| is executed at the first iteration that meets a predefined grasp score threshold (\ref{eq:grasp_score}). This approach is infeasible for (iii) due to its generation of partial models lacking precise surfaces, which complicates grasp pose determination. Consequently, in the case of (iii), we resort to a predetermined flip via an external manipulator at a specific iteration, assuming accurate post-flip object pose knowledge to ensure that the generated models are of the highest quality.\looseness=-1

Evaluation metrics employed include 1) \textbf{PSNR} (Peak Signal-to-Noise Ratio) for assessing visual fidelity, with validation sets of 64 images per object, and 2) \textbf{F-score}~\cite{knapitsch2017tanks} for measuring geometric accuracy, using point clouds derived from the trained NeRF models via Marching Cubes~\cite{lorensen1998marching}.

\subsection{Other Implementation Details}
In our experimental setup, the cost function parameters, including $\lambda$ and $\alpha_i$, play a pivotal role (see ~\ref{eq:loss_defn} and ~\ref{eq:action_cost_defn}). For our purposes, $\lambda$ is fixed at 1, and the $\alpha$ values are determined based on the relative average durations of their corresponding actions executed by the robot, reflecting a practical consideration of action cost in terms of time. However, the observations translate to other values of hyper-parameters as well.  

Our methodology is implemented on the Kaolin-Wisp framework~\cite{takikawa2022kaolin}, utilizing the InstantNGP model~\cite{muller2022instant}. We train an ensemble of five models on a single NVIDIA A40 GPU. For a given pose, we consider the prediction of the ensemble as the mean of the predictions of individual models. On average, training a single NeRF model takes approximately 36 seconds, while determining the next best action requires about 3 minutes. These processes are amenable to parallelization, potentially reducing computation times significantly. The models are trained with images of $800 \times 800$ resolution, and PSNR evaluations are conducted using images at their full resolutions.
\section{Results}
\begin{table*}[t]
\captionsetup{font=small}
\caption{Ablation results for uncertainty estimation technique. Epi, Total stands for epistemic and overall uncertainty as proposed by \cite{sunderhauf2023density} (with modifications stated in section \ref{sec:ablations}). F-score* represents the F-score values multiplied by 10}
\setlength{\abovecaptionskip}{5pt}
\setlength{\belowcaptionskip}{5pt}
\begin{subtable}{\textwidth}
\centering
\captionsetup{font=footnotesize}
\bgroup
\setlength\tabcolsep{3.5pt}
\begin{tabular}{@{}lcccccccccccc@{}}
\toprule
\multicolumn{1}{l}{\multirow{2}[2]{*}{Uncertainty}} & \multicolumn{2}{c}{Basket} & \multicolumn{2}{c}{Cheezit Box} & \multicolumn{2}{c}{Mug} & \multicolumn{2}{c}{Rubik's Cube} & \multicolumn{2}{c}{Spam Can} & \multicolumn{2}{c}{\textbf{Total}} \\
\cmidrule(lr){2-11}\cmidrule(lr){12-13}  & PSNR$\uparrow$  & F-score*$\uparrow$ & PSNR$\uparrow$  & F-score*$\uparrow$ & PSNR$\uparrow$  & F-score*$\uparrow$ & PSNR$\uparrow$  & F-score*$\uparrow$ & PSNR$\uparrow$  & F-score*$\uparrow$ & PSNR$\uparrow$  & F-score*$\uparrow$ \\
\midrule
\multicolumn{13}{c}{Model quality after 20 iterations \textbf{without} grasping}\\
\midrule
Ours   & \textbf{\val{17.2}{0.1}} & \textbf{\val{4.2}{0.5}} & \textbf{\val{21.8}{0.2}} & \textbf{\val{4.3}{0.2}} & \textbf{\val{26.3}{0.1}} & \textbf{\val{6.5}{0.2}} & \textbf{\val{30.3}{0.3}} & \val{3.8}{0.1} & \textbf{\val{23.9}{0.6}} & \textbf{\val{4.1}{0.3}} & \textbf{\val{23.9}{0.1}} & \textbf{\val{4.6}{0.1}} \\
Entropy~\cite{lee2022uncertainty} & \val{14.2}{0.3} & \val{3.0}{0.2} & \val{19.8}{0.2} & \val{3.4}{0.1} & \val{25.0}{0.6} & \val{5.6}{0.2} & \val{27.7}{0.6} & \val{4.0}{0.4} & \val{20.3}{0.3} & \val{4.0}{0.2} & \val{21.4}{0.2} & \val{4.0}{0.1} \\
Epi~\cite{sunderhauf2023density} & \val{15.5}{0.6} & \val{3.2}{0.3} & \val{20.4}{0.3} & \val{3.7}{0.1} & \val{25.8}{0.3} & \val{6.0}{0.4} & \val{29.0}{0.1} & \textbf{\val{4.5}{0.7}} & \val{21.0}{0.4} & \val{3.8}{0.1} & \val{22.3}{0.2} & \val{4.2}{0.2} \\
Total~\cite{sunderhauf2023density} & \val{16.6}{0.4} & \val{3.5}{0.2} & \val{19.7}{0.6} & \val{3.8}{0.3} & \val{21.2}{2.2} & \val{4.1}{0.7} & \val{25.6}{0.5} & \val{3.0}{0.2} & \val{22.1}{1.0} & \val{3.9}{0.1} & \val{21.0}{0.5} & \val{3.7}{0.2} \\
\midrule
\multicolumn{13}{c}{Model quality attained given a cost budget of 2 \textbf{without} grasping}\\
\midrule
Ours   & \textbf{\val{17.1}{0.1}} & \textbf{\val{3.9}{0.3}} & \textbf{\val{21.4}{0.2}} & \textbf{\val{4.3}{0.1}} & \textbf{\val{26.0}{0.3}} & \textbf{\val{5.8}{0.5}} & \textbf{\val{29.8}{0.7}} & \textbf{\val{4.5}{0.2}} & \textbf{\val{23.7}{0.4}} & \textbf{\val{4.1}{0.2}} & \textbf{\val{23.6}{0.2}} & \textbf{\val{4.5}{0.1}} \\
Entropy~\cite{lee2022uncertainty} & \val{14.8}{1.0} & \val{3.0}{0.2} & \val{19.3}{0.5} & \val{3.6}{0.1} & \val{24.9}{0.5} & \val{5.5}{0.1} & \val{27.7}{0.6} & \val{4.0}{0.4} & \val{20.3}{0.3} & \val{4.0}{0.2} & \val{21.4}{0.3} & \val{4.0}{0.1} \\
Epi~\cite{sunderhauf2023density} & \val{15.2}{0.2} & \val{3.1}{0.2} & \val{19.3}{0.2} & \val{3.4}{0.1} & \val{25.2}{0.3} & \val{5.3}{0.3} & \val{28.4}{0.7} & \val{4.5}{0.4} & \val{20.7}{0.2} & \val{4.0}{0.0} & \val{21.8}{0.2} & \val{4.1}{0.1} \\
Total~\cite{sunderhauf2023density} & \val{16.6}{0.4} & \val{3.5}{0.1} & \val{19.7}{0.6} & \val{3.8}{0.3} & \val{21.2}{2.2} & \val{4.1}{0.7} & \val{25.6}{0.5} & \val{3.0}{0.2} & \val{22.1}{1.0} & \val{3.9}{0.1} & \val{21.0}{0.5} & \val{3.7}{0.2} \\

\bottomrule
\end{tabular}%
\egroup
\end{subtable}

\label{tab:metrics_ablations}
\end{table*}

\subsection{Evaluation of Model Quality}
We now showcase the superiority of our active learning pipeline in generating more accurate NeRF models compared to established baselines. 
This section presents our findings under two distinct conditions: (i) \verb|Flip()| actions prohibited, and (ii) \verb|Flip()| actions permitted. First, to ensure an equitable comparison, particularly with ActiveNeRF, which does not focus on minimizing cumulative action costs, we conduct our experiments and those of the baselines across a uniform number of iterations, set at 20. This fixed iteration count is selected to afford ample iterations for all methods to converge to reasonable NeRF models for manipulation.

Subsequently, we investigate the ability of our method to construct higher-quality NeRF models within a predefined total action cost budget, aligning with our original research objective. The allocated cost budget is carefully chosen to be sufficiently generous, enabling the active learning frameworks to develop robust models, yet not so ample as to lead to quality saturation. Specifically, the cost budget is set to 3 for scenarios allowing the \verb|Flip()| action and to 2 for those that do not, with the difference directly correlating to the cost associated with a flip action.

The results are summarized in Table \ref{tab:metrics_experiments}. These results are derived from training both our model and the baselines on segmented images, ensuring a consistent basis for comparison. The proposed method improves PSNR (by 14\%) and F-Score (by 20\%) compared to  ActiveNeRF.

\subsection{Benefit of Object Re-orientation}

\begin{figure}[t!]
    \setlength{\belowcaptionskip}{-10pt}
     \centering
        \includegraphics[width=0.5\textwidth]{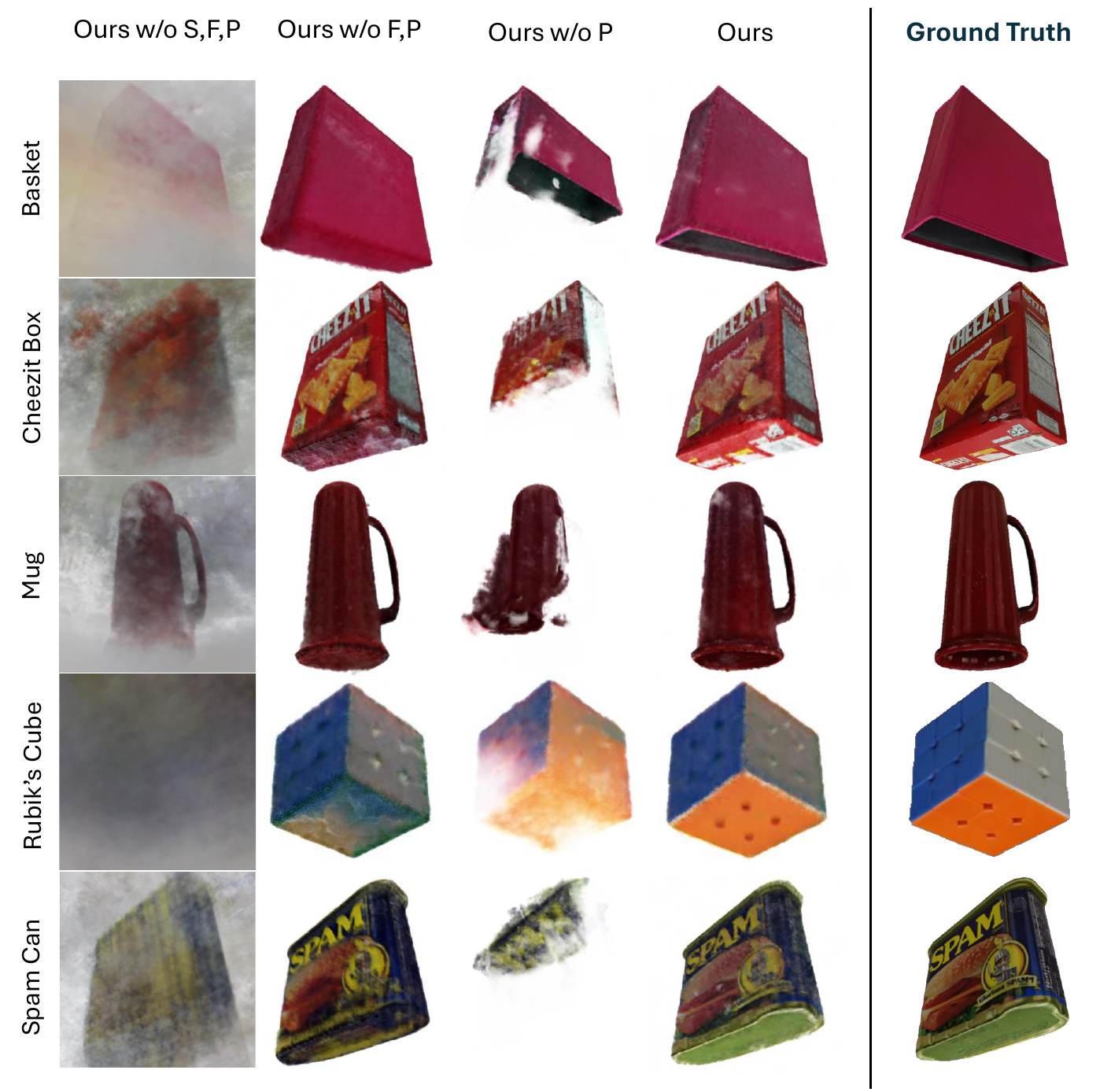}
        \caption{
           \textbf{Model Quality Comparison under Pipeline Modifications.} Comparison of learned model quality with ground truth meshes. The different models are obtained after removing certain components from our pipeline. \textbf{S} denotes \textit{object segmentation}, \textbf{F} denotes \textit{flip}, and \textbf{P} denotes \textit{pose re-acquisition} after the flip has been executed
        }
        \label{fig:qual_res}
\end{figure}

Fig.~\ref{fig:rq2_psnr_iterations} shows that the model quality improves significantly after flipping and exposing previously unseen surfaces. We also show images rendered from models (trained for 20 iterations each) without \verb|Flip()| and compare it against our best model (Fig.~\ref{fig:qual_res}). 
We conclude from the figure that the bottom surface of the object can only be learned by re-orienting the object.
\begin{table}[t!]
\caption{Fraction of Correct Pose Estimation Instances within Bounds \textbf{Tight Bound} := (Rotation Error \textless{} 2\textdegree{} AND Translation Error \textless{} 0.5 cm)   
      \textbf{Loose Bound}:= (Rotation Error \textless{} 5\textdegree{} AND Translation Error \textless{} 1.0 cm) Loss- \textit{Single}: SSD of a single image, \textit{Multi}: summed SSD of 4 images }
  \centering
    \begin{tabular}{@{}l S[table-format=3.2] S[table-format=2.1] S[table-format=3.2] S[table-format=3.2]@{}}
    \toprule
    \multirow{2}[2]{*}{Optimization Method} & \multicolumn{2}{c}{Tight Bound (in \%)} & \multicolumn{2}{c}{Loose Bound (in \%)} \\
    \cmidrule(lr){2-3} \cmidrule(lr){4-5} & {Single} & {Multi} & {Single} & {Multi} \\
    \midrule
    Nelder-Mead \cite{nelder1965simplex} & 17.9   & 33.3   & 30.8   & 59.0 \\
    COBYLA \cite{powell1994direct} 
    \cite{powell1998direct} 
    & 5.1    & 23.1   & 7.7    & 51.3 \\
    Powell \cite{powell1964efficient} & 51.3   & 61.5   & 56.4   & 64.1 \\
    \textbf{Ours (Combined)}   & \textbf{51.3}   & \textbf{69.2}   & \textbf{61.5}   & \textbf{82.1} \\
    \bottomrule
    \end{tabular}%
  \label{tab:inerf_res}%
\end{table}%

\subsection{Grasping Performance Evaluation}
To evaluate the practical utility of the NeRF models, we construct a dataset comprising objects placed in random positions and orientations on a tabletop setup. The robot is tasked to estimate the object pose using a trained NeRF model, followed by an attempt to execute a grasp. The effectiveness of the NeRF models is quantified using GSR (Grasp Success Rate). We conduct a comparative analysis of the performance of NeRF models developed with our active learning methodology and with ActiveNeRF, both with and without the inclusion of the \verb|Flip()| action. The outcomes of this comparison, including the GSR and a breakdown of failure modes for all model variants, are depicted in Fig. \ref{fig:task_based}. We note that ActiveNeRF, is unable to grasp any object, whereas the GSR for our proposed approach is 71\%.

\subsection{Ablations}
\label{sec:ablations}
First, to show the necessity of each component of our pipeline, we remove them step-by-step and run for 20 iterations each. The qualitative results are shown in Fig. \ref{fig:qual_res}.
Next, we delve into the effectiveness of different optimization techniques for pose re-acquisition, including Nelder-Mead~\cite{nelder1965simplex}, COBYLA~\cite{powell1998direct}, and Powell's method~\cite{powell1964efficient}, alongside our approach of selecting the minimum loss among these. The comparative analysis extends to single versus multi-image optimization strategies, as elucidated in Section \ref{sec:reacquisition}, with outcomes presented in Table \ref{tab:inerf_res}. Notably, Fig. \ref{fig:qual_res} demonstrates the crucial role of pose re-acquisition, highlighting that its absence results in significantly degraded NeRF models, rendering them impractical for robotic applications.

Finally, we conduct ablation studies on various uncertainty estimation methodologies. We assess the entropy-based uncertainty metric introduced by Lee et al.~\cite{lee2022uncertainty} and the epistemic and total uncertainties delineated by S{\"u}nderhauf et al.~\cite{sunderhauf2023density}.  As discussed in \ref{sec:model_uncertainty}, their epistemic uncertainty takes on the maximum possible value of $1$ for the pixels lying outside the segmented image of the object.
Since these pixels should not contribute to object uncertainty, we assign them a value of 0.  The results are shown in Table \ref{tab:metrics_ablations}.

\section{Conclusion}
In this paper, we introduced an active learning framework designed to enhance Neural Radiance Fields (NeRF) object models through physical robot interactions, facilitating the revelation of previously occluded surfaces. A notable limitation of our methodology is its computational demand, primarily due to the necessity of ensemble training. Although this process benefits from parallelization, it necessitates multiple GPUs for training with high-resolution images. Furthermore, our pose re-acquisition strategy, despite its general efficacy, occasionally fails to accurately determine the object's pose, as indicated in Table \ref{tab:inerf_res}. Future work will explore improving the timing efficiency of ensembling, extension to articulated objects, considering sequential interactions and experiments on a real manipulation platform. 

\renewcommand*{\bibfont}{\footnotesize}
\printbibliography
\end{document}